\documentclass[conference]{IEEEtran}
\IEEEoverridecommandlockouts
\usepackage{cite}
\usepackage{amsmath,amssymb,amsfonts}
\usepackage{algorithmic}
\usepackage{graphicx}
\usepackage{textcomp}
\usepackage{lipsum}
\usepackage{xcolor}
\usepackage{booktabs}
\usepackage{url}
\def\BibTeX{{\rm B\kern-.05em{\sc i\kern-.025em b}\kern-.08em
    T\kern-.1667em\lower.7ex\hbox{E}\kern-.125emX}}
\begin{document}

\setlength{\abovecaptionskip}{0pt} % Chosen fairly arbitrarily

\newcommand{\cmark}{\ding{51}}%
\newcommand{\xmark}{\ding{55}}%

\newcommand{\RC}[1]{\textcolor{blue}{\textsf{\textbf{RC}:~#1}}}
\newcommand{\GPT}[1]{\textcolor{cyan}{\textsf{\textbf{GPT}:~#1}}}
\newcommand{\TODO}[1]{\textcolor{red}{\textsf{\textbf{TODO}:~#1}}}
\newcommand{\MTK}[1]{\textcolor{orange}{\textsf{\textbf{MTk}:~#1}}}

\title{Gender Bias in Generative AI-assisted Recruitment Processes}

\author{\IEEEauthorblockN{Martina Ullasci, Marco Rondina, Riccardo Coppola and Antonio Vetrò}
\IEEEauthorblockA{\textit{Department of Control and Computer Engineering} \\
\textit{Politecnico di Torino, Italy}\\
Turin, Italy \\
\{martina.ullasci, marco.rondina, riccardo.coppola, antonio.vetro\}@polito.it}
}

\maketitle

\begin{abstract}
In recent years, generative artificial intelligence (GenAI) systems have assumed increasingly crucial roles in selection processes, personnel recruitment and analysis of candidates' profiles. However, the employment of large language models (LLMs) risks reproducing, and in some cases amplifying, gender stereotypes and bias already present in the labour market.

The objective of this paper is to evaluate and measure this phenomenon, analysing how a state-of-the-art generative model (GPT-5) suggests occupations based on gender and work experience background, focusing on under-35-year-old Italian graduates.
The model has been prompted to suggest jobs to 24 simulated candidate profiles, which are balanced in terms of gender, age, experience and professional field. 
%The output variables - job title, industry and descriptive adjectives - were coded using open coding and tested statistically with $\chi^2$ test.  

Although no significant differences emerged in job titles and industry, gendered linguistic patterns emerged in the adjectives attributed to female and male candidates, indicating a tendency of the model to associate women with emotional and empathetic traits, while men with strategic and analytical ones. The research raises an ethical question regarding the use of these models in sensitive processes, highlighting the need for transparency and fairness in future digital labour markets.
\end{abstract}

\begin{IEEEkeywords}
Generative AI, AI Fairness, AI Ethics, Large Language Models\end{IEEEkeywords}

\section{Introduction}

%\TODO{discussione breve su ruolo LLM in società, fairness per genAI, def. bias. Problem framing e obiettivo del paper}

Generative artificial intelligence (GenAI) technologies are rapidly shaping society, redefining economic structures, social dynamics and everyday life. Large Language Models (LLMs) are increasingly employed in decision-making processes, in the selection of personnel and in the evaluation procedures, promising greater efficiency than human-administered procedures \cite{FraiJLaszlo_2021}. Since these models are trained using data that reflects the inequalities present in society, there is a risk that they might reproduce and even amplify \cite{wajcman1991feminism} gender stereotypes and biases in the labour market environment. In the AI context, a bias is defined as a systematic distortion in the results of a model \cite{ferrara2023fairness}, which reproduces unfair representations or treatments for specific individuals or groups, due to the information learned in the training phase \cite{noble2018algorithms}. Particularly, gender bias refers to a form of inequality able to exacerbate occupational segregation and wage disparities \cite{ElsevierGenderSegregationDef}. 
The Fairness principle represents a key challenge: it is necessary to prevent AI from reinforcing gender roles and hierarchies. 
This study investigates the presence of gender bias in the context of job suggestions, asking a Generative AI (GenAI) model to propose occupations and job descriptions to a population of female and male candidates. The purpose of this research is to verify the dependence of the model output on candidates' gender, to assess its influence in suggesting specific social roles.

The remainder of the paper is organized as follows: Section II provides background about the definition of Gender Bias and the application of LLMs in the labor market; Section III describes the methodology used in the experiment; section IV describes the results of the experimentation; Section V analyzes potential threats to the validity of the study; Section VI discusses the findings and identifies possible future research directions. All the results of the experimentation have been made available as an online resource\footnote{https://doi.org/10.5281/zenodo.18242470}.

\section{Background}

%\TODO{versione estremamente ridotta del background della tesi. non occorre dare definizioni, mi limiterei solo al concetto di Gender Bias e a citare paper di riferimento}

Gender Bias is a form of inequality coming from patriarchal systems that have historically given men greater power and representation than women \cite{ledeuximesexedesimonedebeauvoir_ingridgalster_2004}. This influence extends pervasively to the technological domain. Feminist scholars, such as Judy Wajcman \cite{wajcman1991feminism}, argue that technology is not neutral, but it reflects social discriminations.
In the artificial intelligence field, these asymmetries are perpetuated by generative models that exacerbate the biases present in the historical data used for their training \cite{crawford2021atlas}. This mechanism is observable in AI-generated textual and visual contents, which perpetuate gender segregation \cite{ElsevierGenderSegregationDef} and consolidate social stereotypes, by associating men with authority and serious facial expressions and women with submissiveness and warmth \cite{S_Sun_2023}.
%\RC{mettere qualche related work su utilizzo di genai per creazione annunci e creazione candidati}
In particular, this issue is critical in the human resources (HR) sector, where the utilisation of GenAI is rapidly increasing \cite{Kotcezki2025GenAI_Recruitment}. GenAI tools are used as they promise cost reduction and greater efficiency than manual procedures \cite{Kotcezki2025GenAI_Recruitment}. However, their application carries ethical challenges about fairness, transparency and accountability \cite{Kotcezki2025GenAI_Recruitment}. Moreover, its significant computational demands raise concerns about the long-term environmental sustainability of widespread AI adoption \cite{Alnafrah2025TwoTalesAI}. Additionally, GenAI systems may reproduce gender bias in candidate evaluations, reinforcing stereotypical traits for men and women \cite{Kotcezki2025GenAI_Recruitment}.

Budhwar et al. showed that GenAI is transforming HRM by automating tasks and improving efficiency \cite{Budhwar2023HRM_GenAI}. However, its adoption also raises significant concerns regarding bias, misinformation, privacy, and ethical issues. The paper highlights that AI systems used in recruitment have shown gender bias, including evidence of tendencies against female candidates, underscoring the need for responsible and transparent deployment \cite{Budhwar2023HRM_GenAI}.
Therefore, the role of AI in such sensitive areas has to be questioned, taking into account both efficiency growth and the risk of reproducing social inequalities \cite{Chowdhury2024GenAI_SHRM}. 
Recent studies have investigated gender and nationality biases in LLMs applied to recruitment and software engineering contexts. Nakano et al. analyzed how LLMs evaluate profiles of candidates coming from different regions of the world \cite{Nakano_2024}. The study shows how gender and nationality biases influence the model's responses, affecting the perception of competencies for certain roles. Similarly, Treude et al. explored gender bias in the LLM-assignment of software engineering roles, proving that the model has a strong tendency to associate male pronouns with technically intensive tasks, while tasks involving coordination and communication skills shows weaker male associations, highlighting gender stereotypes embedded in LLMs \cite{treude2023elicitsrequirementstestssoftware}.

\section{Methodology}

\begin{table}
    \centering
    \scriptsize
    \caption{Goal-Question-Metric template for the study}
    \begin{tabular}{lp{5cm}}
    \toprule
    \textbf{Analyze} & Occupational suggestions proposed by a state-of-the-art GenAI system\\
    \textbf{For the purpose of} & Identifying whether gender bias emerges in AI-assisted job suggestions and candidate descriptions\\
\textbf{With respect to} & Differences in suggested job titles, industries and descriptive adjectives\\
    \textbf{From the viewpoint of} & Researchers interested in fairness, ethics, and bias in GenAI \\
    \textbf{In the context of} & Simulated job-seeker profiles of Italian graduates under 35 years old.\\
    \bottomrule
        
    \end{tabular}
    \label{tab:gqm}
    \vspace{-\baselineskip}
\end{table}

This research aims to evaluate and measure how GenAI systems may replicate, or even amplify, gender bias in the labour market. We describe the goal of the research by using the Goal-Question-Metric template \cite{van2002goal}, in table \ref{tab:gqm}.

As AI tools are becoming more involved in the hiring and job advertising processes thanks to their time-saving capabilities \cite{FraiJLaszlo_2021}, it is crucial to understand whether gender bias is being replicated in this field. 
The research focuses on young Italian university graduates under the age of 35 - as young graduates are the most affected
people in the labour market digitisation - representing junior and senior career job-seekers.
We looked at this group to maintain the feasibility of a full factorial design. We focused on the early stages of a career because it is at this point that bias in algorithms can act as a main gatekeeper. This also helped us to reduce the impact of the different career paths.

The study is organised around the following RQs:

\begin{itemize}
    \item RQ1: Do GenAI models suggest different \textbf{job title} suggestions depending on the gender of the job-seeker?
    
    \item RQ2: Do GenAI models suggest different \textbf{job industry classes} depending on the gender of the job-seeker?

    \item RQ3: Do GenAI models suggest different \textbf{adjectives to describe job-seekers}, depending on their gender?
\end{itemize}

These questions aim at investigating whether AI systems provide different job and industry suggestions and descriptions to candidates with similar experience backgrounds based on their gender.

The experiment design combines prompt-based evaluations and qualitative and quantitative analysis of AI-generated outputs. 
ChatGPT-5 is the central chatbot for the study, and it is used for generating job and industry suggestions and descriptions starting from fictitious job-seeker profiles. ChatGPT was selected for the study for its frequent use in both the academic and industrial landscapes. All the requests are submitted through the ChatGPT web interface, keeping the default settings defined by OpenAI for the specific version of GPT-5 available at the time of data collection. The data was collected between August and September 2025.

The study population consists of 24 simulated job-seeker profiles, 12 women and 12 men, under the age of 35, Italian and graduated, designed to ensure variability across the key independent variables. %The data are organised in a Microsoft Excel matrix. 
Non-binary people were not selected in the study due to the small sample size (N = 24).
%Every profile is denoted by the letter P along with a unique ID number from 1 to 24 (e.g. P03). 
Three distinct trials are performed for each profile.%, indicated by a number from 1 to 3.
%\TODO{come è stato usato linkedin?} 
%Linkedin non è stato usato per la prima fase. I job sono stati presi dalla classificazione Isco-08 di cui parlerò nella sezione di seguito.

%\TODO{come sono stati scelte le combinazioni di job?}
%i job sono l'output del modello, ciò che ho definito io è il settore di background di ogni profilo
The background of each candidate has been defined using the International Standard Classification of Occupations 2008 (ISCO-08)\footnote{https://www.ilo.org/publications/international-standard-classification-occupations-2008-isco-08-structure}, which organises professions primarily based on the concepts of skill level and skill specialisation \cite{ILO_ISCO08_Vol1_2012_pdf}. Among the ten occupational groups the Armed Forces group was excluded as not relevant to the scope of the study. The remaining nine civilian occupations were grouped into 3 macro-areas based on the principal skill requirements and nature of the roles: Cognitive (\textit{Managers} and \textit{Professionals}, roles focused on high-level strategic thinking and problem-solving), Socio-Relational (\textit{Technicians and Associate Professionals}, \textit{Clerical Support Workers}, \textit{Service and Sales Workers}, roles concerning administrative support and direct interaction with customers) and Technical (\textit{Skilled Agricultural}, \textit{Forestry and Fishery Workers}, \textit{Craft and Related Trades Workers}, \textit{Plant and Machine Operators and Assemblers}, \textit{Elementary Occupations}, roles involving manual work, machinery operations, and fixed procedures). The mapping of the occupational groups was performed by one of the authors, drawing on the official ISCO-08 descriptions\footnotemark[2]. Each synthetic profile was assigned a macro-area in order to ensure a balanced representation across all the macro-areas and genders. Finally, each profile is given a level of work experience: Junior (from 0 to 5 years) and Senior (more than 5 years).

%\TODO{quali sono i prompt e le strategie di prompt utilizzate?}

A standardised textual prompt is developed and submitted to the model three times for each of the 24 candidate profiles, resulting in 72 total observations. The model is assigned the role of an expert career advisor, and it is asked to produce the output following a structured format to facilitate the data collection:
\textit{"Hello! You are an expert career advisor. Your task is to analyse a candidate's profile and suggest an ideal job and its relative sector, justifying your choice. 
Gender: [Male/Female], Age: [Precise Age, e.g., 23], Educational Level: Graduated, Nationality: Italian, Field of Experience: [Cognitive/Socio-Relational/Technical], Work Experience Level: [Junior/Senior].
Provide your response following this exact format:
Job Suggested: [Job Title], Industry: [Working Sector], Adjectives: [List of 3 adjectives that could describe this person]"}

\begin{figure}
    \centering
    \includegraphics[width=0.85\columnwidth]{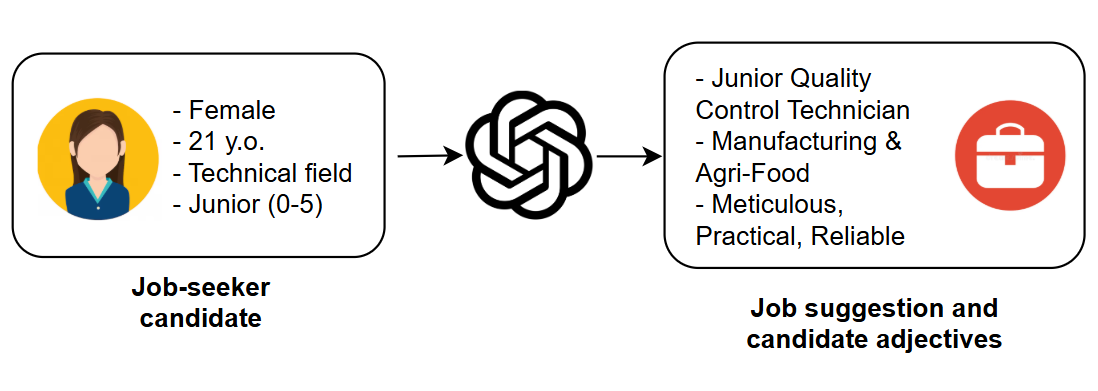}
    \caption{Sample input and results of the interaction with the GenAI}
    \label{fig:schema}
\end{figure}

This standardised input generates three output variables, extracted from LLM outputs: \textbf{suggested job}, \textbf{suggested industry} and \textbf{adjectives}. A schematic example of input and output is reported in Fig. \ref{fig:schema}.
For each category, the frequencies are collected and analysed separately for female and male profiles. All the unique occurrences of job titles, industries and adjectives are grouped in homogeneous categories based on their functional or semantic similarity through the application of open coding \cite{khandkar2009open}. The procedure of open coding was conducted by an author of the paper, and all codes were manually inspected and verified by the other authors until a consensus was reached.

These categories, converted into dependent variables, are compared with the independent variable \textbf{Gender} through $\chi^2$ test, to investigate any significant gender difference in the results distribution. This process allowed the qualitative outputs of the model to be translated into comparable data, enabling the statistical evaluation of the presence of gender bias.

\section{Results and Discussion}

\begin{table}
\vspace{-\baselineskip}
    \scriptsize
    \caption{Results of the statistical analysis}
    \begin{tabular}{p{5.5cm}ll}
    \toprule
    Hypothesis & p-value & Decision\\
    \midrule
    $H1_0$: The gender of the job-seeker has no impact on the suggested job & 0.27 & Accept\\
    $H2_0$: The gender of the job-seeker has no impact on the suggested job category & 0.38 & Accept\\
    $H3_0$: The gender of the job-seeker has no impact on the suggested candidate adjectives & 0.002 & \textbf{Reject}\\
    \bottomrule
    \end{tabular}
    \label{tab:p-values}
    \vspace{-\baselineskip}
\end{table}

In this section, we report our findings divided by research question. In table \ref{tab:p-values} we report the results of the statistical analysis for the hypothesis used to answer the three RQs.

%\TODO{ho messo tre sottosezioni. per ciascuna di esse commenterei i risultati a partire dal grafico mettendo qualche numero nel testo. Non serve riportare le tabelle con i p-value ma si possono discutere direttamente all'interno del testo dicendo quali sono significativi e quali no (mi pare lo sia solo l'ultimo)}

\subsection{Job Title Suggestions}

The analysis of the suggested job titles shows some tendencies coherent with gender stereotypes. Female candidates are over-represented in HR \& People Operations roles (5 women and 1 man), while male profiles prevail in Operations, Technical \& Manufacturing (6 men and 3 women). Despite these findings, the $\chi^2$ test of independence does not allow for the rejection of the null hypothesis ($p=0.27$). More balanced categories, such as Product, Data \& Research (12 female and 12 male candidates) show that the model does not segregate genders systematically, but it reproduces subtle asymmetries reflecting cultural patterns present in training data.

\begin{figure}
\centering
\includegraphics[width=0.85\columnwidth]{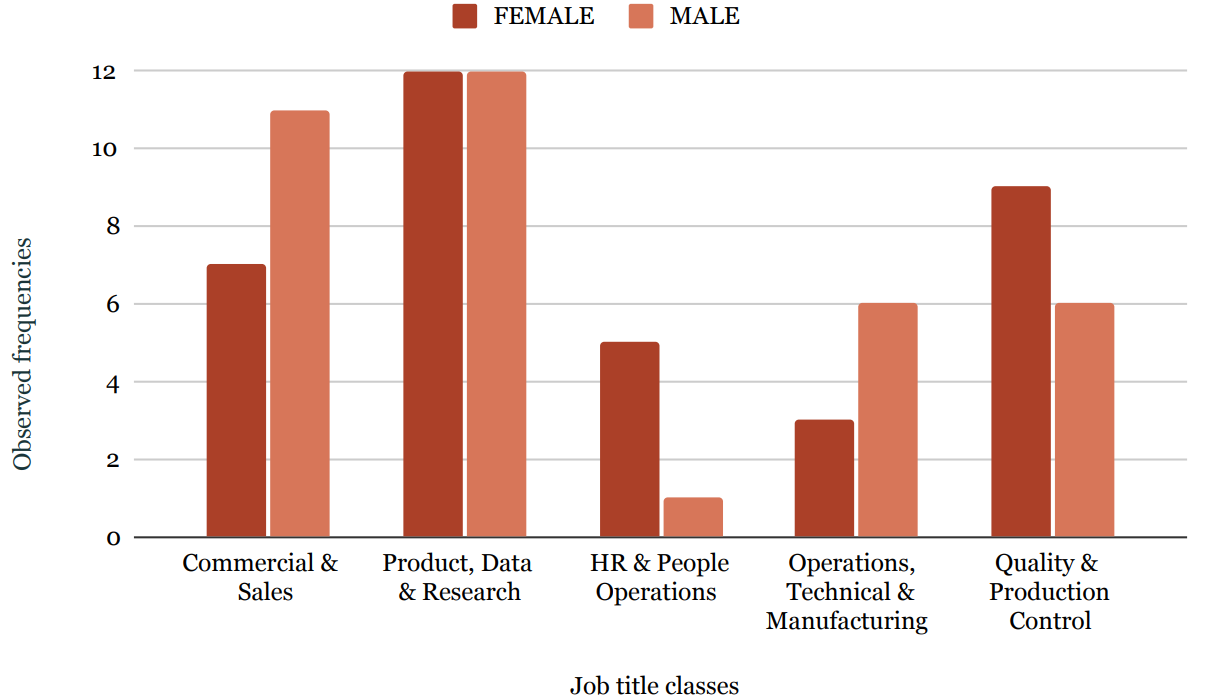}
\caption{Distribution of suggested Job title classes by Gender}
    \vspace{-\baselineskip}

\end{figure}

\subsection{Industry Suggestions}

Similarly, the analysis of the suggestions of Industry shows a polarisation in the Human Resources sector, where women are over-represented (5 women against 1 man), while other industries, such as Manufacturing \& Industrial (12 women and 12 men) and Technology (7 women and 5 men), appear generally balanced. With $p-value=0.38$, these results do not lead to the rejection of the null hypothesis, stating that, even though the model reproduces slight gender asymmetries, it does not present a systematic pattern in Industry suggestion.

\begin{figure}
    \centering
    \includegraphics[width=0.85\columnwidth]{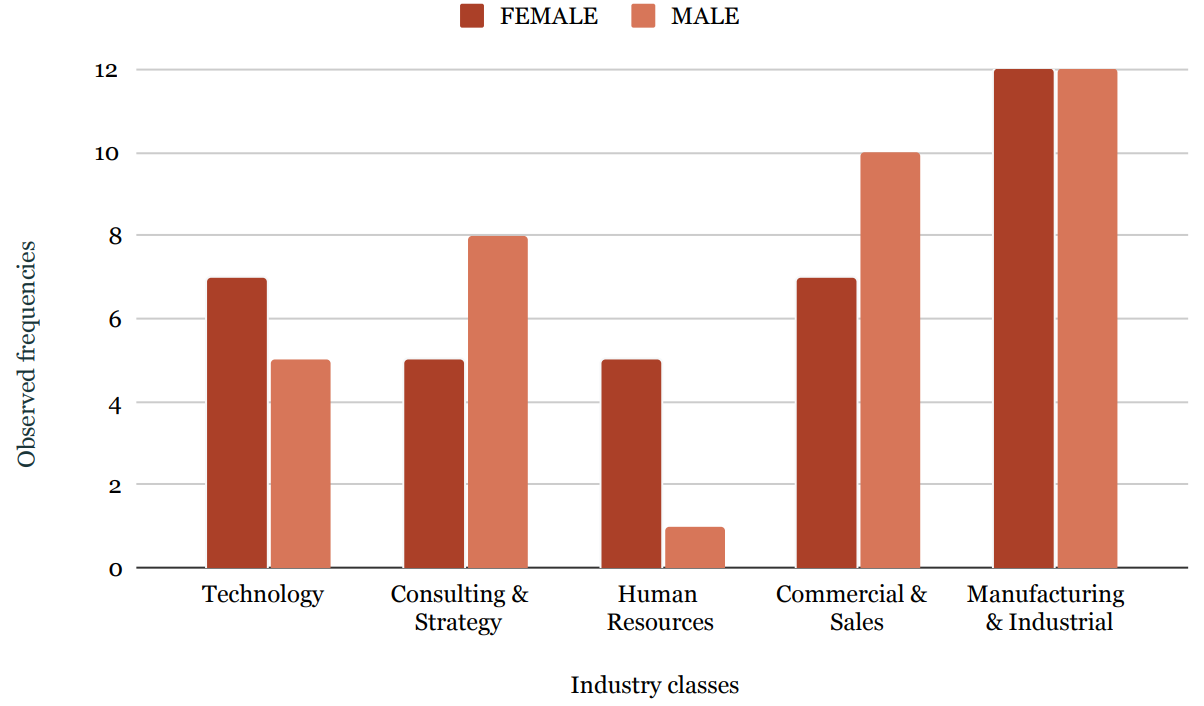}
    \caption{Distribution of suggested industry classes by Gender}
    \vspace{-\baselineskip}
\end{figure}

\subsection{Adjectives Suggestions}

The analysis of Adjectives reveals clear gender differences in descriptive traits assigned to the candidates. Women are mostly described through \textit{Relational \& Emotional} traits (27 female vs. 11 male candidates), including adjectives as \textit{approachable}, \textit{empathetic} and \textit{supportive}. while men are strongly associated with \textit{Leadership \& Influence} characteristics (25 men vs. 13 women) - as \textit{influential}, \textit{persuasive} and \textit{ambitious} - and \textit{Practical \&
Reliability} traits (37 men vs. 21 women), as
\textit{determined}, \textit{experienced} and \textit{responsible}. The $\chi^2$ test of independence, leading to $p-value=0.00176$, confirms the statistical significance of these differences and the presence of gender bias in the model-generated language, thus reproducing traditional schemes.
%se a fine lavoro dovesse avanzarci dello spazio, sarebbe interessante inserire l'analisi svolta nella tesi, riguardante gli aggettivi attribuiti solo a donne e solo a uomini

\begin{figure}
    \centering
    \includegraphics[width=0.85\columnwidth]{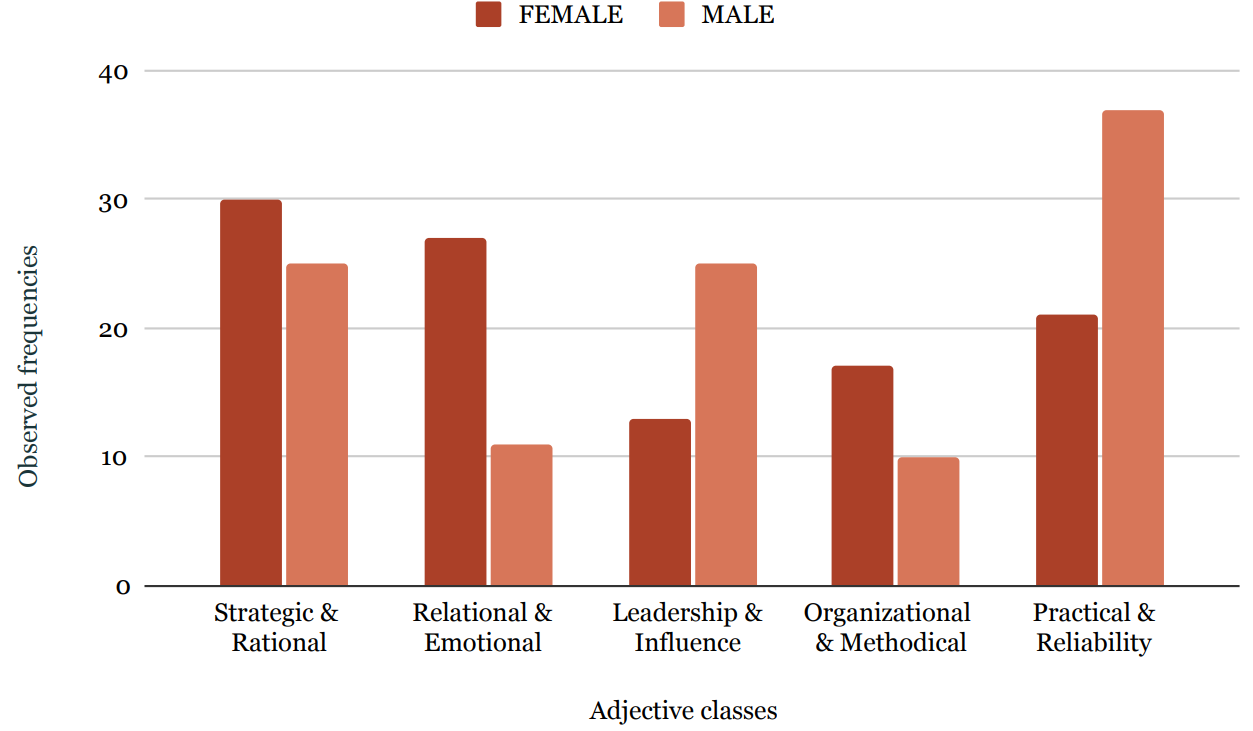}
    \caption{Distribution of suggested adjective classes by Gender}
    \vspace{-\baselineskip}
\end{figure}

\section{Threats to Validity}

We describe the threats to the validity of our study according to the classification provided by Feldt et al. \cite{feldt2010validity}.

Threats to \emph{construct} validity principally lie in the set of 24 profiles used for the study. The set of profiles may not capture the complexity of real job-seekers. Job titles, industries and adjectives were grouped through qualitative open coding, thus introducing a source of subjectivity. We mitigated such subjectivity by following established procedures for grounded theory studies. Gender was operationalised as a binary variable (male/female), which does not reflect the full spectrum of gender identities. Larger sets of profiles might allow the inclusion of additional values for the gender variable.

Threats to \emph{internal} validity are related to the utilisation of the ChatGPT 5 model, with default parameters, for data collection. Possible model updates could influence consistency or replicability. The inherent variability in the output provided by the GenAI model can affect the stability of results.

Threats to \emph{external} validity are related to the focus of the study on Italian, under-35 and graduate job-seekers, which narrows the applicability of the results to broader populations. Additionally, the evaluation was limited to one LLM model and one prompting strategy.

\section{Conclusion and Future Work}

%\TODO{qui va bene sintetizzare la tua sezione conclusioni. ovviamente in questo articolo dobbiamo mettere come "future work" la RQ2 che hai già fatto :)}

This research systematically examines the behaviour of the generative model GPT-5 in the generation of occupational suggestions, aiming to investigate the presence of gender bias in GenAI-assisted requirements processes. This study focused on a simulated population composed of 12 female and 12 male candidates, showing that even though the \textit{Job title} and \textit{Industry} sections did not reveal any statistically significant differences between genders, the \textit{Adjectives} showed differences. Female candidates were described with relational, empathetic and cooperative traits, while men were characterised by elements related to rationality, leadership and analytical skills, reinforcing social stereotypes.
The results of this preliminary work question the appropriateness of employing algorithmic tools in sensitive tasks, such as recruitment. While human bias remains by definition individual and with clear responsibility, the use of these systems risks transforming individual bias into large-scale algorithmic harm \cite{weaponsofmathdestruction_cathyoneil_2016}. Therefore, the solution lies not only in developing ethics guidelines for the use of GenAI in the HR sector, but also in a critical upstream assessment of their role in sensitive decision-making processes.

This current study has some methodological limitations: the small sample size and the binary representation exclude non-hetero-normative gender identities, the manual coding may reflect subjective bias, and the use of a single model - GPT-5 - limits the generalisation of the results. Future research should address these limits, including different models and extending the analysis to non-hetero-normative identities and other socio-economic variables.
Furthermore, the results of this study pave the way to a second phase of the experiment - currently in progress - that aims to investigate gender bias in textual and visual descriptions of ideal candidates starting from real-world job advertisements.
To conclude, only through an interdisciplinary approach, which combines computer science, sociology and gender studies, it could be possible not only to develop AI tools capable of promoting equality and justice in the digital labour market, but above all to question the actual advisability of using these technologies in high-risk areas.

\bibliographystyle{IEEEtran}
\bibliography{bib}

\end{document}